\documentclass[conference]{IEEEtran}
\IEEEoverridecommandlockouts
\usepackage{cite}
\usepackage{url}

\usepackage{mathtools}
\usepackage{amsmath,amssymb,amsfonts}
\DeclareMathOperator{\E}{\mathbb{E}}

\usepackage{algorithmic}

\usepackage{graphicx}
\graphicspath{{figures/}}
\usepackage{caption}
\usepackage{subcaption}
\usepackage{float}
\usepackage{textcomp}
\usepackage{xcolor}
\def\BibTeX{{\rm B\kern-.05em{\sc i\kern-.025em b}\kern-.08em
    T\kern-.1667em\lower.7ex\hbox{E}\kern-.125emX}}
\begin{document}

\title{Stellar Cluster Detection using GMM with Deep Variational Autoencoder\\
}

\author{\IEEEauthorblockN{Arnab Karmakar}
\IEEEauthorblockA{\textit{Department of Avionics}\\\textit{Indian Institute of Space Science}\\\textit{and Technology, Trivandrum} \\
arnab.sc15b079@ug.iist.ac.in}
\and
\IEEEauthorblockN{Deepak Mishra}
\IEEEauthorblockA{\textit{Department of Avionics}\\\textit{Indian Institute of Space Science}\\\textit{and Technology, Trivandrum} \\
deepak.mishra@iist.ac.in}
\and
\IEEEauthorblockN{Anandmayee Tej}
\IEEEauthorblockA{\textit{Department of Earth ans Space Science}\\\textit{Indian Institute of Space Science}\\\textit{and Technology, Trivandrum} \\
tej@iist.ac.in}
}

\maketitle

\begin{abstract}
Detecting stellar clusters have always been an important research problem in Astronomy. Although images do not convey very detailed information in detecting stellar density enhancements, we attempt to understand if new machine learning techniques can reveal patterns that would assist in drawing better inferences from the available image data. This paper describes an unsupervised approach in detecting star clusters using Deep Variational Autoencoder combined with a Gaussian Mixture Model. We show that our method works significantly well in comparison with state-of-the-art detection algorithm in recognizing a variety of star clusters even in the presence of noise and distortion.
\end{abstract}


\section{Introduction}
Studies of stellar clusters are crucial in developing an understanding of the universe. These are ideal laboratories for astrophysical research. Clusters contain statistically significant populations of stars spanning a wide range of stellar mass within a relatively small volume of space. Since these stars share the common heritage of being formed more or less simultaneously from the same progenitor molecular cloud, observations of cluster have been used to unveil crucial aspects of stellar evolution theory. Detecting these clusters in the background of densely populated sky is difficult due to the presence of molecular clouds, unrelated background objects and instrument noise.\par

Application of Machine Learning has not yet well explored in the domain of Astronomy due to the unavailability of structured (and labelled) dataset. George et al.\cite{ml1} applied Deep Learning in detecting gravitational waves, whereas Viquar et al.\cite{ml2} applied SVM-KNN and AdaBoost in Quasar-Star classification. A comprehensive overview of Machine Learning algorithms in Astronomy has been given by Ball\cite{ml3}. Recently, the application of Generative Adversarial Networks in recovering features from degraded astronomical images has been a significant contribution (Schawinski\cite{ml4}). To the best of our knowledge, Unsupervised Learning specifically Variation Autoencoder based detection model have not been previously tried in stellar cluster detection.

Detection of star clusters is not new. Initially clusters were detected manually\cite{b1}, though recent studies mostly use the density enhancement on background field. Kumar et al.\cite{kumar} used star counts in spatial binning to construct the surface density maps. Schmeja et al.\cite{schmeja} give a comprehensive overview on the cluster detection methods and implemented four algorithms, namely star counts\cite{kumar}\cite{lada1995}\cite{sc3}, nearest neighbour method\cite{nn1}\cite{nn2}, Voronoi tessellation\cite{vt1}\cite{vt2}, and the separation of minimum spanning trees\cite{b5}\cite{mst2}\cite{mst3}. Although these methods hold good, a lot of manual tuning has to be used before arriving at a satisfactory result.\par

The earlier work, as mentioned before, uses catalogue data (i.e. source points, refer \ref{data_format}) considering individual stellar sources for verifying the reliability of their cluster detection algorithms. For a given astronomical data (image and source points) detecting a stellar cluster is still a challenging task and in this work we attempt for the first time implementation of deep learning for the same. We propose an unsupervised approach in detecting the cluster from image data. We divide the image into patches and represent a binary classification problem, where positive label of the patch represent that the patch belongs to a particular cluster.

We implement a Deep Variational Autoencoder (DeepVAE) (inspired from Kingma and Welling, 2014\cite{vae}), where the latent representation from the inference network is coupled with a Gaussian Mixture Model (GMM) for producing a better estimation of cluster area. The assumption that a star cluster follow a Gaussian distribution is fairly accurate for globular clusters, and we model the cluster and the background as 2 separate Gaussian. Our model defines latent variables to be drawn from a Mixture-of-Gaussian (MoG) for better separability in GMM clustering. We use Stochastic Gradient Variational Bayes (SGVB) with reparameterization trick to minimize the negative evidence lower bound (NELBO). The results of our detection framework hold good with the astronomically detected standard clusters. Our contribution focuses on the application of deep unsupervised learning in the field of stellar cluster detection from astronomical images.\par

\section{Data Collection}

\begin{figure*}[h]
\centering
\begin{subfigure}{.5\textwidth}
  \centering
  \includegraphics[width=0.98\linewidth]{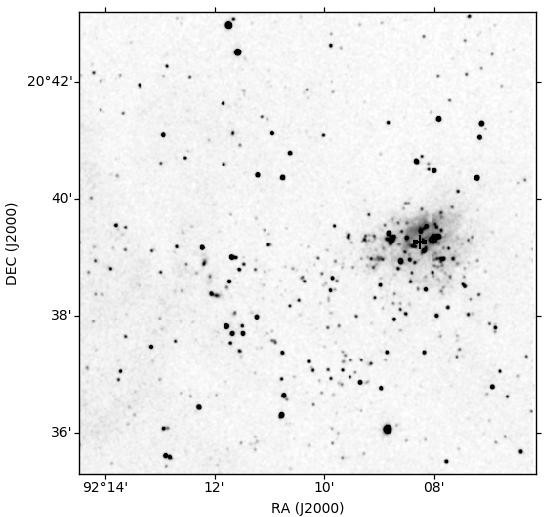}
  \label{fig:sub1}
\end{subfigure}%
\begin{subfigure}{.5\textwidth}
  \centering
  \includegraphics[width=0.98\linewidth]{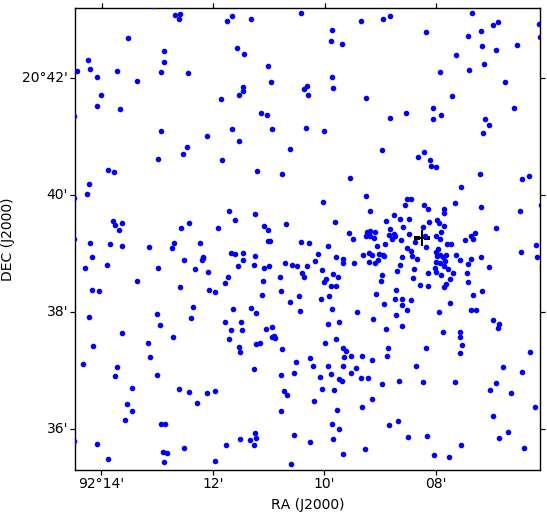}
  \label{fig:sub2}
\end{subfigure}
\caption{The 2MASS $K_s$-band image (left) and the corresponding point source plot (right) of the region around IRAS 06055+2039 object. The plus sign marks the position of the IRAS point source. Image as well as catalogue data is plotted in terms of astronomical coordinates (Right Ascension: RA and Declination: DEC, in degrees; equivalent to X and Y coordinates) for consistency.}
\label{fig:data_demo}
\end{figure*}

\subsection{Near Infrared Data from UKIDSS}
We use the K-band (2.20$\mu$m) data from UKIRT Infrared Deep Sky Survey (UKIDSS) 10PLUS Galactic Plane Survey (GPS) archive\footnote{\url{http://wsa.roe.ac.uk/}}, which has a resolution of $\sim$ 1 arcsec and 5$\sigma$ limiting magnitude
of K = 18.1 mag\cite{ukidss1}.
\subsection{Near Infrared Data from 2MASS}
NIR ($K_s$ band) data for point sources as well as images around any given cluster position can be obtained from the Two Micron All Sky Survey (2MASS)\cite{2mass1} Point Source Catalog (PSC) and 2MASS image service archive\footnote{\url{https://irsa.ipac.caltech.edu/Missions/2mass.html}} respectively.
\subsection{Data format}\label{data_format}
\subsubsection{Image Data}
Images from both the archives are obtained as 16-bit single channel FITS image files. We have used only images for detection with the proposed DeepVAE model. For accuracy assessment we have overlaid the detection results over detections from catalogued data for consistency.
\subsubsection{Catalogue Data}
Catalogues are a form of storing each individual astronomical objects as point source with specific properties. Each source has its defined position in terms of astronomical coordinates with corresponding magnitudes(brightness), uncertainties and other parameters. We have filtered them so that only stars or probable stars are present in the retrieved dataset (UKIDSS: \verb|mergedClass| values of -1 or -2 \cite{ukidss2}; 2MASS: \verb|read-flag| values 1 to 6 \cite{2mass1}).

\section{GMM-DeepVAE model}
The image is divided into patches, which is the input $x$ to our DeepVAE model, and we extract the rich feature embeddings $z$ from the network for GMM clustering.
For a variational autoencoder, the joint distribution over observed variable $x$ and latent variable $z$ is defined as,
\begin{equation}
p_\theta(x,z) = p_\theta(x|z)p(z)
\end{equation}
Here we assume that $z$ is drawn from a prior Mixture of Gaussian (MoG) density $p(z)$ and related with the observed variable with the likelihood $p_\theta(x|z)$.

\begin{figure*}
 	\centering
	\includegraphics[width=\textwidth]{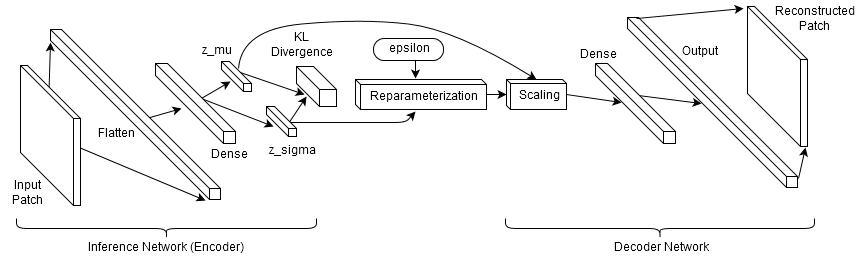}
	\caption{Representation of the DeepVAE model. The Dense layer is the representation of a fully connected network of 3 hidden layers. We encapsulating the divergence in an auxiliary layer for modularity. We extract z\_mu and z\_sigma from this framework to use for our GMM clustering }
\label{fig:deepvae}
\end{figure*}

\subsection{Decoder}
Our primary assumption is that $p_\theta(x|z)$ follows a multivariate Bernoulli with $3$ fully connected layers,
\begin{align}
p_\theta(x|z) &= Bern(\sigma(WH+b))
\end{align}
where $\sigma$ is logistic sigmoid function, $H$ is previous layer output and $\theta = \{W, b\}$ are the model parameters.\par
The decoder part of the network takes the (reparameterized) latent variable $z$ as input and learns to reconstruct $\tilde{x}$.

\subsection{Inference network}
For each observed variable $x_n$, we approximate the posterior $p_\theta(z_n|x_n)$ using a variational distribution $q_{\phi_n}(z_n|x_n)$ with local variational parameters $\phi_n = \{\mu_n,\sigma_n \}$ as
\begin{equation}\label{gauss}
q_\phi(z_n|x_n) = \mathcal{N}(z_n|\mu_\phi(x_n),diag(\sigma_\phi^2(x_n)))
\end{equation}
where $\mu_\phi(x_n)$ and $\sigma_\phi(x_n)$ are the model estimate to $\mu_n$, $\sigma_n$. Finally, we optimize our network to learn global variational parameters $\phi$ instead of learning $\phi_n$ for each data point.\par
We derive the mean $\mu_\phi$ and log variance $\log(\sigma_\phi^2)$ as the output of out inference network consisting fully connected neural network with 3 hidden layers.

\subsection{Loss function}
The optimum variational parameter $\phi$ is obtained by minimizing the Kullback-Leibler (KL) divergence of approximate posterior $q_\phi(z|x)$ to the marginal likelihood $p_\theta(x)$
\begin{equation}\label{KL}
\phi^* = argmin_\phi KL[q_\phi(z|x)||p_\theta(z|x)]
\end{equation}
Both $p_\theta(x)$ and the KL divergence suffers intractability. Therefore we maximize the alternative divergence, the evidence lower bound (ELBO) expressed as
\begin{align}\label{elbo}
ELBO(q) &= \E_{q_\phi(z|x)}[\log p_\theta(x|z) + \log p(z) - \log q_\phi(z|x)]\nonumber\\
		&= \E_{q_\phi(z|x)}[\log p_\theta(x|z)] - KL[q_\phi(z|x)||p(z)]
\end{align}
Maximizing $ELBO(q)$ with model parameter $\theta$ and variational parameter $\phi$ approximately maximizes log marginal likelihood $p_\theta(x)$ as well as minimizes KL divergence (equation \ref{KL}) respectively.\par
The first term in this loss function (equation \ref{elbo}) is the reconstruction term, and the KL divergence regularizes the latent embeddings $z$ to lie in the Mixture of Gaussian (MoG) manifold.\par
In this work instead of maximizing the ELBO, we have used the negative evidence lower bound (or NELBO) directly as the loss function to minimize using Stochastic Gradient Variational Bayes (SGVB) estimator.\par
The KL term (equation \ref{elbo}) can actually be calculated using a closed form expression (Kingma and Welling, 2014\cite{vae})
\begin{equation}
KL[q_\phi(z|x)||p(z)] = -\frac{1}{2}\sum_{k=1}^K \{ 1 + \log\sigma_k^2 - \mu_k^2 - \sigma_k^2 \}
\end{equation}

\subsection{Reparameterization}
The gradient of ELBO (or NELBO) is intractable due to its dependency in both $\theta$ and $\phi$. Through reparameterization, the gradient of ELBO is expressed as an expectation of the gradient, then we use the Stochastic Gradient Descent (SGD) on the repeated Monte Carlo (MC) gradient estimates.\par
It is a simple change of variable expressing $z \sim q_\phi(z|x)$ as a deterministic transform $z \sim g_\phi(x,\epsilon)$, where $\epsilon \sim p(\epsilon)$.
Writing ELBO as an expectation of function $f(x,z)$ and substitute $z$ as above, we get
\begin{align}
\nabla_\phi\E_{q_\phi(z|x)}[f(x,z)] &=  \nabla_\phi\E_{p(\epsilon)}[f(x,g_\phi(x,\epsilon))] \nonumber \\
&= \E_{p(\epsilon)}[\nabla_\phi f(x,g_\phi(x,\epsilon))]
\end{align}
For the diagonal gaussian approximation (equation \ref{gauss}), we introduce a location-scale transformation
\begin{equation}
z = g_\phi(x,\epsilon) = \mu_\phi(x) + \sigma_\phi(x) \boldsymbol{\cdot} \epsilon, \quad \epsilon \sim \mathcal{N}(0,I)
\end{equation}

\subsection{Gaussian Mixture Model (GMM)}
The prior assumption for $p(z)$ being a MoG allows us to model the actual cluster and the background to be generated from two different Gaussian models. Therefore, we perform a binary classification (clustering) where we predict whether the image patch belong to cluster or background.

We use the feature rich embeddings $z$ as input and optimize the GMM using the traditional Expectation-Maximization (EM) algorithm \cite{gmm}. The basic intuition is discussed here. 

\noindent{\rule{\linewidth}{0.1pt}}
{\flushleft{\textbf{Initialization}}\par}
randomly initialize class means $\mu_c$ and covariance $\Sigma_c$\par
initialize class probabilities $p(c)=\frac{1}{2}$ \par
{\flushleft{\textbf{E-step}}\par}
calculate probability of class c (defined as a Gaussian distribution), responsible for $z_i$
{\flushleft{\textbf{M-step}}\par}
update $\mu_c$, $\Sigma_c$, $p(c)$
{\flushleft{\textbf{Reinitialize}}\par}
\textbf{if} $p(c)$ is low \textbf{or} very similar $\mu_c$
{\flushleft{\textbf{Check for convergence}}\par}
{\flushleft{\textbf{Classification}}\par}
\textbf{for} each point $z$ find $c$ that maximizes $p(c|x)$
\noindent{\rule{\linewidth}{0.1pt}}

\section{Training}
\begin{figure}[h]
  \centering
  \includegraphics[width=0.7\linewidth]{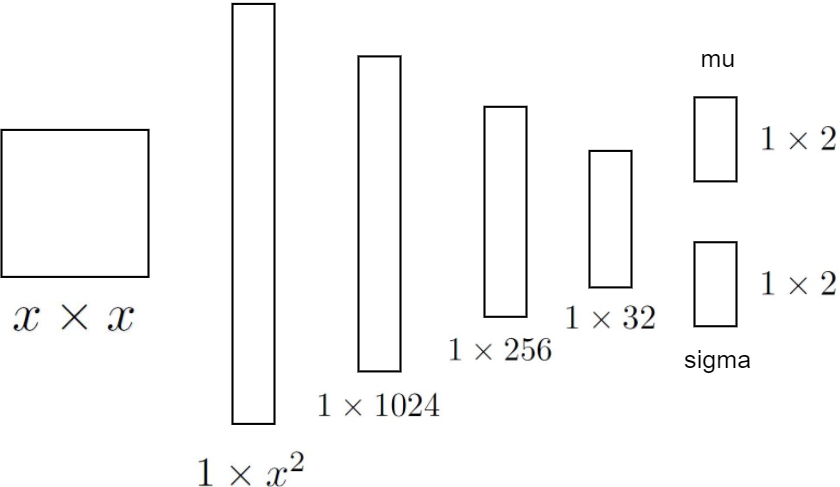}
  \caption{The detailed Dense layer network structure. Encoder extracts $z_\mu$ and $z_\sigma$ whereas decoder converts reparameterized $\epsilon_\mu$ and $\epsilon_\sigma$ to reconstructed image patch.}
  \label{fig:training}
\end{figure}

We divide the image into patches of $8\times8$, $16\times16$, $32\times32$ and $64\times64$ with 50\% overlap (e.g. extracting $16\times16$ patches with stride of 8). This generates a set of 4 datasets with different sample sizes from one image. Unique patch sized datasets have been trained independently with our DeepVAE model. The encoder and decoder follow similar structure with 3 hidden layers with ($1024, 256, 32$ nodes respectively). The last layer is also fully connected with the $\mu$ and $\sigma$ layer. The Dense layer in Fig. \ref{fig:deepvae} is detailed in Fig. \ref{fig:training}.

\begin{figure}
  \centering
  \includegraphics[width=\linewidth]{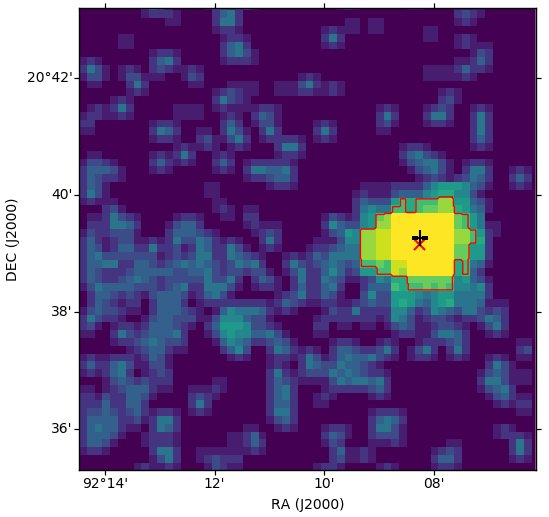}
  \caption{The heatmap plot combining detections from all 4 different patch sizes. We use an adaptive threshold ($0.7\times\text{maximum value}$) to detect the final cluster area (red).}
  \label{fig:heatmap}
\end{figure}

The GMM classifier predicts binary class for each individual patch using the extracted $z_\mu$ and $z_\sigma$ values. From the classification results, we backtrace and highlight the patches predicted as positive by our network \ref{fig:heatmap}.

We use the ensemble of the results from all 4 datasets to reconstruct a heatmap for identifying the maximally detected region. After experimenting with different clusters (table \ref{tab1}), we finalize the adaptive threshold of $0.7\times\text{(maximum value)}$ to filter out false detections and get the final segmented out region. This we predict as the cluster region in the image.

\section{Results and Discussion}
\begin{figure}
 	\centering
	\includegraphics[width=\linewidth]{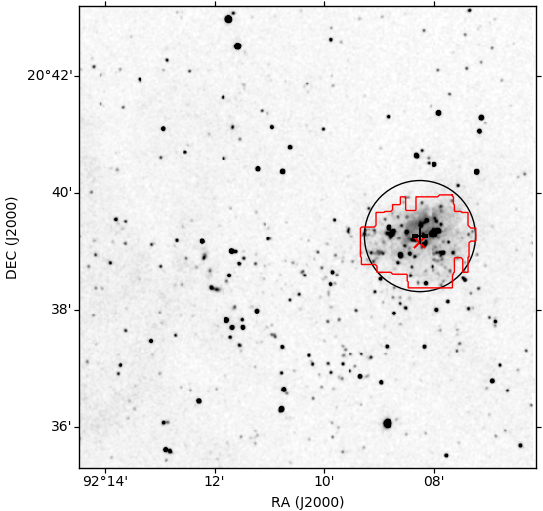}
	\caption{Comparison of detected cluster using our algorithm (GMM-DeepVAE in red) and detection results from Tej. et al.\cite{tej-1}(black circle) around IRAS 06055+2039. The red cross marks the detected center according to our algorithm and the plus sign marks the position of the IRAS point source.}
	\label{fig:detection}
\end{figure}

From the final detection map, we infer the center of the cluster as a weighted average of pixel position and intensity. Also, we define the cluster area as number of pixels scaled by astronomical coordinates ($1\times1$ pixel $\sim$ $0.2\times0.2$ arcsec for UKIDSS and $1\times1$ pixel $\sim$ $1.03\times1.03$ arcsec for 2MASS datasets). The radius of the cluster is inferred by modelling the detected area (in pixels) as a circle and computing its radius.
For accuracy assessment, we superimpose the detected region on point source plot and use the Intersection-over-Union (IoU) as the parameter calculated on source count.
\begin{equation}
\text{IoU} = \frac{\text{objects (point sources) in the area of Intersection}}{\text{objects (point sources) in the area of Union}} \nonumber
\end{equation}
Obtained IoU for IRAS 06055+2039 is 85.95\%. Since our model precisely extracts the region belonging to significant stars and then we model it as a circle, the prediction for radius is almost always slightly less that the state-of-the-art values. We found the radius of IRAS 06055+2039 as $77.13"$ compared to $85"$ and number of member stars as $89$ compared to $98$ by Tej. et al.\cite{tej-1}. Final detection results for are shown in Fig. \ref{fig:detection}.

\begin{figure}
 	\centering
	\includegraphics[width=0.65\linewidth]{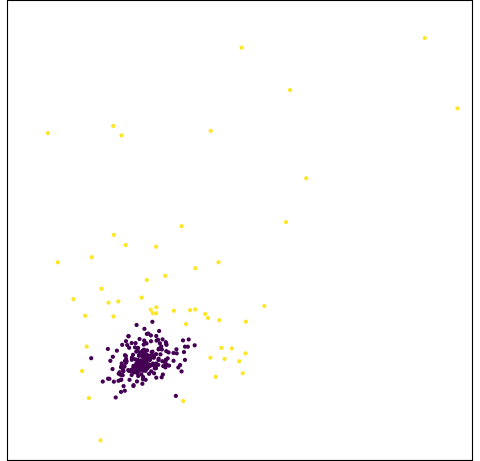}
	\caption{Visualization of latent embeddings $z$ as classified by GMM for $64\times64$ patch size. The background patches are highly clustered with similar properties due to noise or faint stars whereas the actual cluster areas have weak association and diverse population.}
	\label{fig:gmm}
\end{figure}

\begin{table}[htbp]
\caption{Comparison of Results}
\begin{center}
\begin{tabular}{|c|c|c|c|c|c|}
\hline
\textbf{IRAS}&\multicolumn{2}{|c|}{\textbf{Radius ($"$)}}&\multicolumn{2}{|c|}{\textbf{Members}}&\textbf{IoU} \\
\cline{2-5} 
\textbf{Name} & \textbf{\textit{our}}& \textbf{\textit{sota}}& \textbf{\textit{our}} & \textbf{\textit{sota}} & \textbf{\textit{\%}} \\
\hline
06055+2039 & 77.13 & 89 & 98 & 119 & 85.95\\
\hline
05274+3345 & 89.86 & 103 & 56 & 48 & 79.68\\
\hline
05345+3157 & 101.63 & 126 & 98  & 95 & 83.27 \\
\hline
05358+3543 & 116.74 & 130 & 71 & 53 & 74.32 \\
\hline
05490+2658 & 98.46 & 122 & 93 & 95 & 82.24 \\
\hline
05553+1631 & 123.59 & 104 & 96 & 80 & 71.43 \\
\hline
06056+2131 & 104.79 & 132 & 101 & 132 & 78.23 \\
\hline
06061+2151 & 116.05 & 145 & 102 & 105 & 69.98 \\
\hline
\end{tabular}
\label{tab1}
\end{center}
\end{table}

The comparison of various parameters detected by our model with the state-of-the-art (sota) results\cite{tej-1}\cite{kumar} has been shown in TABLE \ref{tab1}. We infer that our results are fairly accurate, considering or model uses only image data without incorporating any other astronomical property of the sources.

\section{Conclusion}
In this paper, we implemented a GMM based classification with DeepVAE for star cluster detection. Our method follows a multi-field-of-view approach as we validate our results with 4 different patch size based training without incorporating any supervised information. We compared the performance of our model with the baselines on multiple clusters and found the results reasonably equivalent.

The next step is combining the more accurate point source data (with multiple parameters) with the image to produce highly accurate results. We use a MoG prior for DeepVAE in this work, although experimentation with alternative divergences can be done, which will be our future work. Also, joint optimization of DeepVAE with GMM can potentially improve the results.

Almost all existing methods use the catalogue data for detection of clusters. Our deep learning based approach on images is significantly faster, without any manual intervention and fairly accurate in comparison. Although images does not contribute much detail about the sources, for some cases like embedded clusters it gives information about molecular cloud overdensity. In cases where only image data is available without individual source labelling, our method can perform well. This can be extended to the more significant problem of galaxy classification, where detecting the morphology of the galaxy gives significant cues of stellar evolution.



\begin{thebibliography}{00}
\bibitem{ml1}
George, Daniel, and E. A. Huerta. "Deep neural networks to enable real-time multimessenger astrophysics." Physical Review D 97.4 (2018): 044039.

\bibitem{ml2}
Viquar, Mohammed, et al. "Machine Learning in Astronomy: A Case Study in Quasar-Star Classification." arXiv preprint arXiv:1804.05051 (2018).

\bibitem{ml3}
Ball, Nicholas M., and Robert J. Brunner. "Data mining and machine learning in astronomy." International Journal of Modern Physics D 19.07 (2010): 1049-1106.

\bibitem{ml4}
Schawinski, Kevin, et al. "Generative adversarial networks recover features in astrophysical images of galaxies beyond the deconvolution limit." Monthly Notices of the Royal Astronomical Society: Letters 467.1 (2017): L110-L114.

\bibitem{b1}  
Messier, C.: 1774, Tables des Nebuleuses, ainsi que des amas d'Etoiles, que l'on decouvre parmi les Etoiles fixes sur l'horizon de Paris; observes a l'Observatoire de la Marine. Memoires de l'Academie des Sciences for 1771, Paris

\bibitem{vae}
Kingma, Diederik P., and Max Welling. "Auto-encoding variational bayes." arXiv preprint arXiv:1312.6114 (2013).

\bibitem{kumar}
Kumar, M. S. N., E. Keto, and E. Clerkin. "The youngest stellar clusters: Clusters associated with massive protostellar candidates." Astronomy \& Astrophysics 449.3 (2006): 1033-1041.
  
\bibitem{schmeja}
Schmeja, S. "Identifying star clusters in a field: A comparison of different algorithms." Astronomische Nachrichten 332.2 (2011): 172-184.

\bibitem{b5} 
Dib, Sami, Stefan Schmeja, and Richard J. Parker. "Structure and mass segregation in Galactic stellar clusters." \textit{Monthly Notices of the Royal Astronomical Society} 473.1 (2017): 849-859.

\bibitem{lada1995}
Lada, Elizabeth A., and Charles J. Lada. "Near-infrared images of IC 348 and the luminosity functions of young embedded star clusters." The Astronomical Journal 109 (1995): 1682-1696.

\bibitem{sc3}
Kirsanova, M. S., et al. "Star formation around the H ii region Sh2-235." Monthly Notices of the Royal Astronomical Society 388.2 (2008): 729-736.

\bibitem{nn1}
Gutermuth, Robert A., et al. "The Spitzer Gould belt survey of large nearby interstellar clouds: discovery of a dense embedded cluster in the Serpens-Aquila Rift." The Astrophysical Journal Letters 673.2 (2008): L151.

\bibitem{nn2}
Gouliermis, Dimitrios A., et al. "Hierarchical stellar structures in the local group dwarf galaxy NGC 6822." The Astrophysical Journal 725.2 (2010): 1717.

\bibitem{vt1}
Van Breukelen, C., et al. "Galaxy clusters at 0.6<z<1.4 in the UKIDSS ultra deep survey early data release." Monthly Notices of the Royal Astronomical Society: Letters 373.1 (2006): L26-L30.

\bibitem{vt2}
Espinoza, Pablo, Fernando J. Selman, and Jorge Melnick. "The massive star initial mass function of the Arches cluster." Astronomy \& Astrophysics 501.2 (2009): 563-583.


\bibitem{mst2}
Maschberger, Th, et al. "Properties of hierarchically forming star clusters." Monthly Notices of the Royal Astronomical Society 404.2 (2010): 1061-1080.

\bibitem{mst3}
Campana, Riccardo, et al. "A Minimal Spanning Tree algorithm for source detection in $\gamma$-ray images." Monthly Notices of the Royal Astronomical Society 383.3 (2008): 1166-1174.


\bibitem{gmm}
Dempster, Arthur P., Nan M. Laird, and Donald B. Rubin. "Maximum likelihood from incomplete data via the EM algorithm." Journal of the royal statistical society. Series B (methodological) (1977): 1-38.

\bibitem{tej-1}
Tej, Anandmayee, et al. "A multiwavelength study of the massive star-forming region IRAS 06055+ 2039 (RAFGL 5179)." Astronomy \& Astrophysics 452.1 (2006): 203-215.

\bibitem{ukidss1}
Dye, Simon, et al. "The UKIRT infrared deep sky survey early data release." Monthly Notices of the Royal Astronomical Society 372.3 (2006): 1227-1252.

\bibitem{ukidss2}
Lucas, P. W., et al. "The UKIDSS Galactic plane survey." Monthly Notices of the Royal Astronomical Society 391.1 (2008): 136-163.

\bibitem{2mass1}
Skrutskie, M. F., et al. "The two micron all sky survey (2MASS)." The Astronomical Journal 131.2 (2006): 1163.


\end{thebibliography}
\end{document}